\documentclass[letterpaper, 10 pt, conference]{ieeeconf}

\usepackage{epsfig}
\usepackage{cite}
\usepackage{booktabs}
\usepackage{multirow}
\usepackage{mathtools}
\usepackage{bm}
\usepackage{algorithm}
\usepackage{comment}
\usepackage{color}
\usepackage{setspace}

\newcommand{\figlab}[1]{\label{fig:#1}}
\newcommand{\figref}[1]{Fig.~\ref{fig:#1}} 
\newcommand{\tablab}[1]{\label{tab:#1}}
\newcommand{\tabref}[1]{Table~\ref{tab:#1}} 
\newcommand{\forlab}[1]{\label{for:#1}}

\newcommand{\etal}{\textit{et~al.}}
\definecolor{green}{rgb}{0.01, 0.5, 0.01}

\usepackage{amssymb}
\usepackage{pifont}

\usepackage{subcaption}
\usepackage{url}
\usepackage{threeparttable}

\usepackage{amsmath}
\usepackage[noend]{algorithmic}
\usepackage{array}

\begin{document}
\title{Soft Regrasping Tool Inspired by Jamming Gripper}
\author{Takuya Kiyokawa$^{1}$ Zhengtao Hu$^{2}$, Weiwei Wan$^{1}$, and Kensuke Harada$^{1,3}$, 
\thanks{$^{1}$Department of Systems Innovation, Graduate School of Engineering Science, The University of Osaka, 1-3 Machikaneyama, Toyonaka, Osaka, Japan.}%
\thanks{$^{2}$School of Mechatronic Engineering and Automation, Shanghai University, Shanghai, China.}%
\thanks{$^{3}$Industrial Cyber-physical Systems Research Center, The National Institute of Advanced Industrial Science and Technology (AIST), 2-3-26 Aomi, Koto-ku, Tokyo, Japan.}%
}

\maketitle

\vspace{-20pt}
\begin{abstract}
Regrasping on fixtures is a promising approach to reduce pose uncertainty in robotic assembly, but conventional rigid fixtures lack adaptability and require dedicated designs for each part. 
To overcome this limitation, we propose a soft jig inspired by the jamming transition phenomenon, which can be continuously deformed to accommodate diverse object geometries. 
By pressing a triangular-pyramid-shaped tool into the membrane and evacuating the enclosed air, a stable cavity is formed as a placement space. 
We further optimize the stamping depth to balance placement stability and gripper accessibility. 
In soft-jig-based regrasping, the key challenge lies in optimizing the cavity size to achieve precise dropping; once the part is reliably placed, subsequent grasping can be performed with reduced uncertainty. 
Accordingly, we conducted drop experiments on ten mechanical parts of varying shapes, which achieved placement success rates exceeding 80\% for most objects and above 90\% for cylindrical ones, while failures were mainly caused by geometric constraints and membrane properties. 
These results demonstrate that the proposed jig enables general-purpose, accurate, and repeatable regrasping, while also clarifying its current limitations and future potential as a practical alternative to rigid fixtures in assembly automation.
\end{abstract}

\IEEEpeerreviewmaketitle

\section{Introduction}
In recent years, variant–variety production has been increasingly demanded in response to market needs for rapid and flexible adaptation to fluctuations in both product types and quantities. To realize such production systems, robotic assembly must be capable of handling diverse parts with both generality and precision.

A major challenge in robotic assembly is grasp uncertainty, which can significantly reduce positioning accuracy in subsequent insertion or alignment tasks. Regrasping, in which an initially grasped part is placed, realigned, and then grasped again, has been shown to be effective in mitigating such uncertainty~\cite{Wan2017}. However, most existing approaches rely on rigid jigs with fixed geometries~\cite{Hu2024}, which are impractical for production lines handling short-lived and diverse products. This limitation motivates the need for regrasping tools that can adapt their cavities flexibly to different part shapes.

To address this need, we propose a flexible jig inspired by the jamming gripper. As shown in~\figref{overview}(a), the jig consists of glass beads enclosed by a silicone membrane, which is fixed around its perimeter by a circular rigid ring. The membrane is easily deformable, as illustrated in~\figref{overview}(b), allowing for the generation of reconfigurable cavities. Specifically, by pressing a triangular-pyramid-shaped part into the membrane and then solidifying the structure, as shown in~\figref{overview}(c), cavities of various triangular-pyramid sizes can be created. This mechanism provides a versatile means of generating adaptable cavities for regrasping tasks.

In this study, we extend existing regrasp planning approaches by incorporating both stability analysis of the final placement pose~\cite{Tsuji2009} and the feasibility of grasp and trajectory generation~\cite{Wan2017}. 
We then evaluate the proposed jig through experiments focusing on the accuracy of triangular-pyramid cavity generation and object drop trials, since once dropping is performed reliably, the subsequent regrasp can be achieved with reduced uncertainty. 
Through tests on ten objects of varying shapes, we also illustrate representative success and failure cases to clarify the method’s limitations and future directions.
\begin{figure}[tb]
  \begin{minipage}[tb]{0.62\linewidth}
    \centering
    \includegraphics[width=\linewidth]{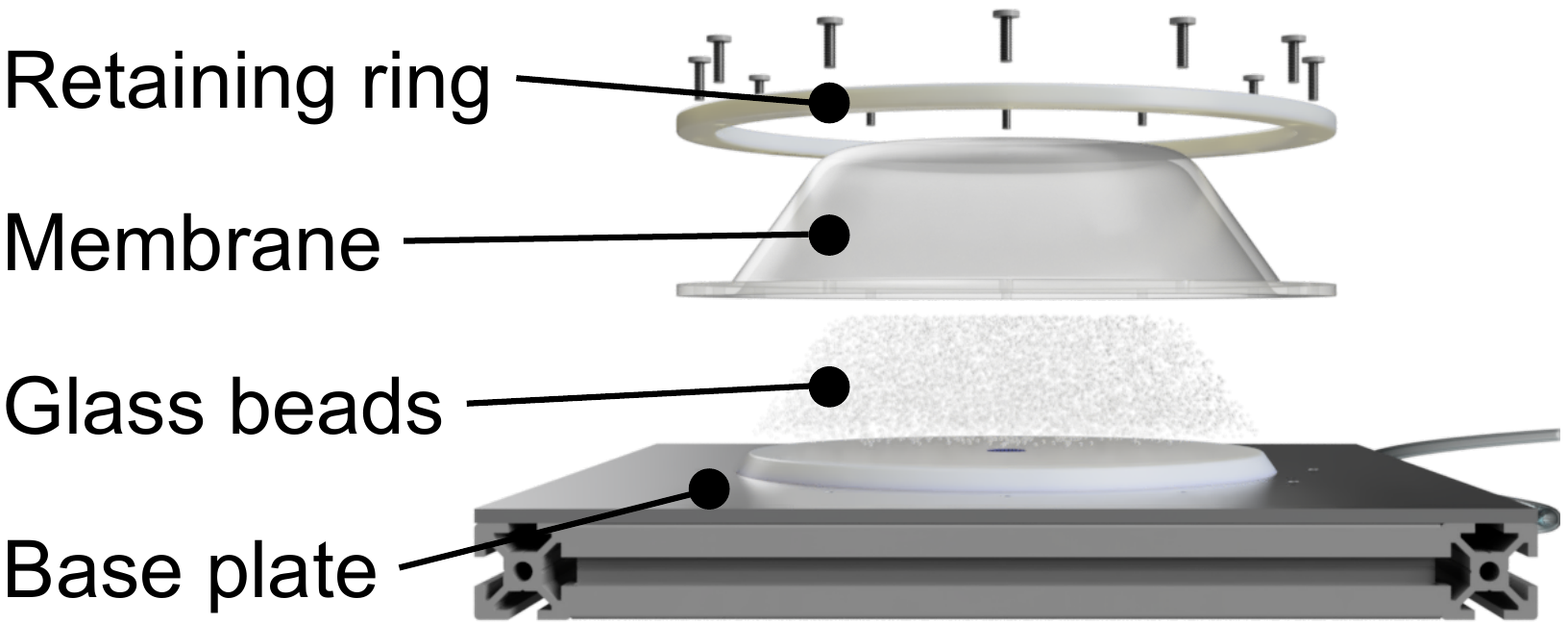}
    \subcaption{Structure}
  \end{minipage}
  \begin{minipage}[tb]{0.36\linewidth}
    \centering
    \includegraphics[width=0.9\linewidth]{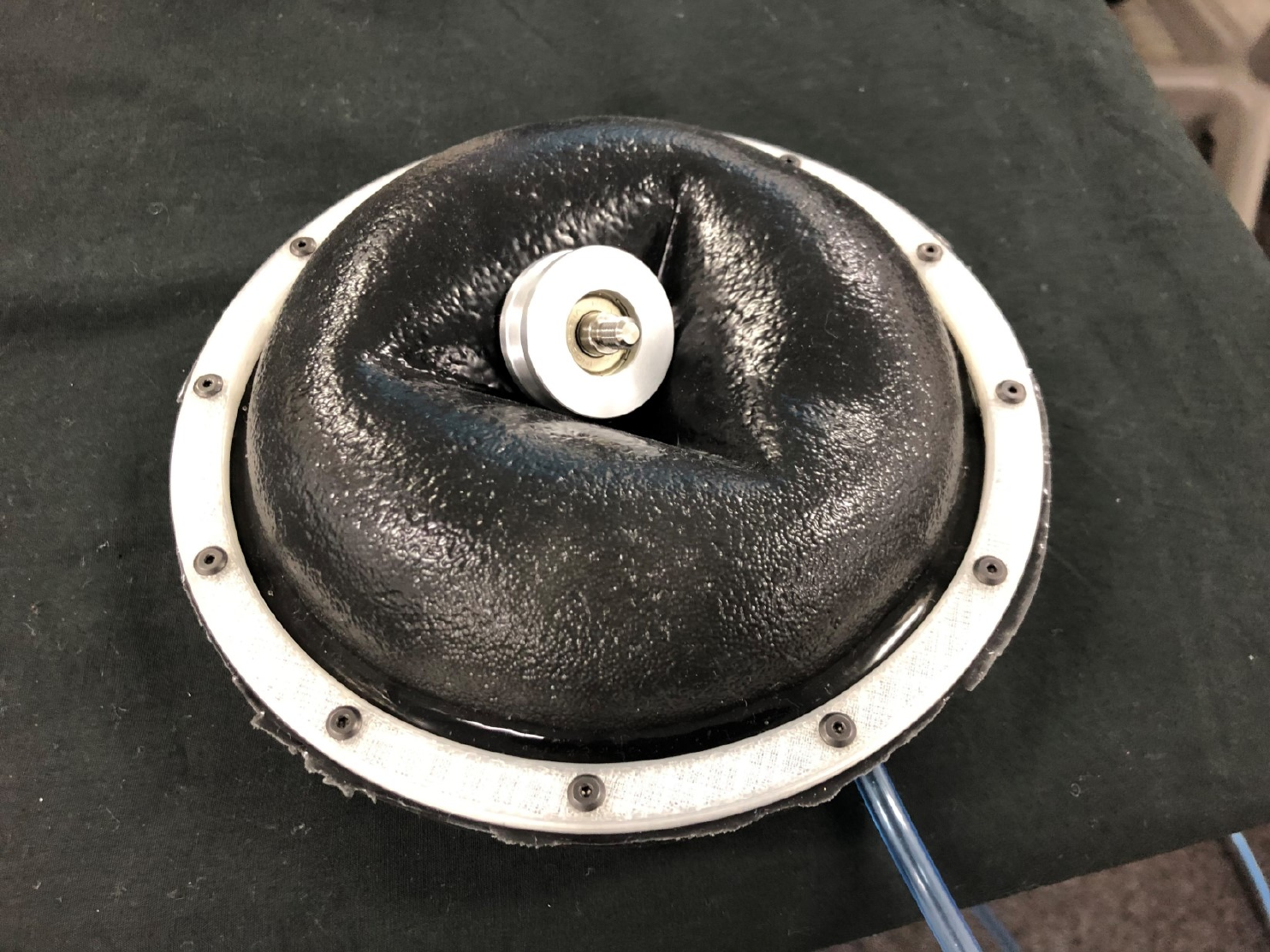}
    \subcaption{Appearance}
  \end{minipage}
  \begin{minipage}[tb]{\linewidth}
    \centering
    \includegraphics[width=\linewidth]{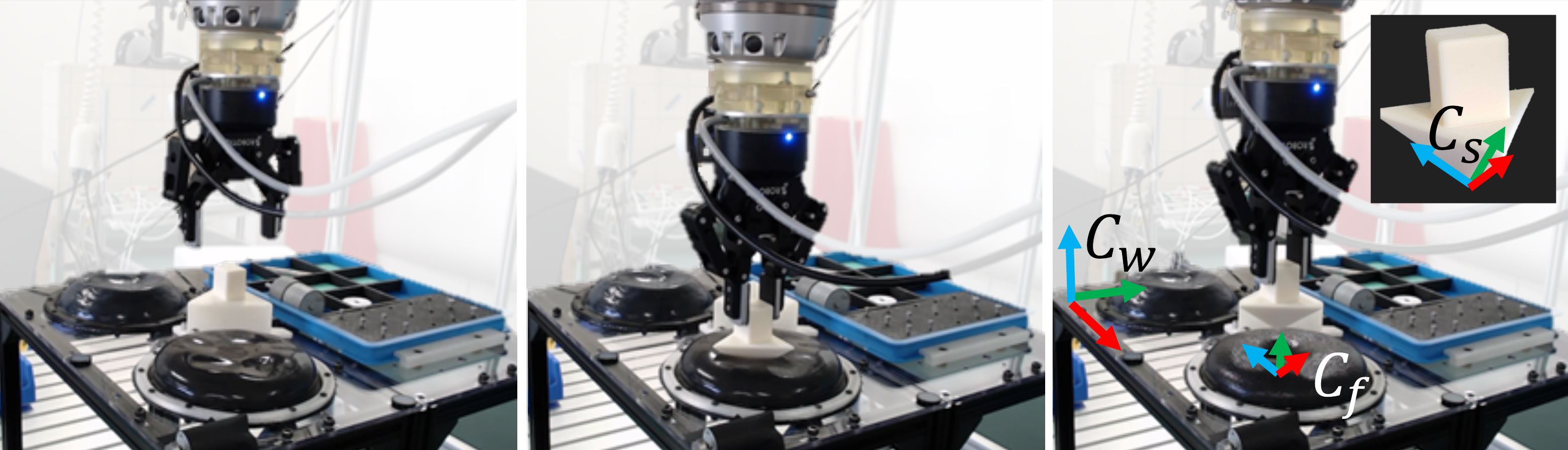}
    \subcaption{Generating a triangular-pyramid cavity}
  \end{minipage}
  \caption{Jamming-gripper-inspired soft fixture} \figlab{overview}
\end{figure}

\section{Related Work}
\subsection{Regrasping-Based Uncertainty Reduction}
Regrasping has been widely studied as an effective means to reduce pose uncertainty and improve success rates in robotic assembly.
Early work explored the use of rigid supports: Hu~\etal~\cite{Hu2024} analyzed regrasping with rigid jigs, showing that pose uncertainty can be reduced by exploiting stable contacts with predefined geometries.
In parallel, Roa~\etal~\cite{Roa2009} proposed a quasi-static regrasp planning approach in the grasp space using independent contact regions, guaranteeing force-closure during finger relocation.

Building on such foundational concepts, more recent approaches have leveraged perception and learning.
Bauza~\etal~\cite{Bauza2024} developed \textit{simPLE}, a visuotactile framework that learns in simulation to integrate task-aware grasping, tactile-based pose estimation, and graph-based regrasp planning.
Wada~\etal~\cite{Wada2022} introduced \textit{ReorientBot}, which combines learned waypoint selection with traditional motion planning to achieve dynamic, single-step reorientation and dual-arm regrasping for specific-posed placement.

A complementary direction has focused on extrinsic actions that exploit the environment and gravity to reduce uncertainty.
Drigalski~\etal~\cite{Drigalski2022} proposed an uncertainty-aware manipulation planner that uses actions such as place, grasp, and push against the environment to minimize pose uncertainty without dedicated jigs.

Beyond industrial pick-and-place, researchers have extended regrasping to domain-specific and complex tasks.
Chiu~\etal~\cite{Chiu2021} addressed bimanual regrasping of suture needles, leveraging reinforcement learning in an ego-centric state space for rapid dual-arm trajectory generation in surgical settings.
Qin~\etal~\cite{Qin2023} further applied dual-arm regrasping to the manipulation of long deformable belts, proposing a hierarchical planner integrated with loco-manipulation to enable successive regrasps in constrained industrial environments.

Together, these studies highlight the potential of regrasping to enhance flexibility and robustness in assembly tasks, while often relying on kinematic or task-specific assumptions that limit generality.

\subsection{Jig-Based Uncertainty Reduction}
In parallel, extensive research has focused on jig and fixture design to enable stable holding and precise positioning of components.
Grippo~\etal~\cite{Grippo1987} and Whybrew~\etal~\cite{Whybrew1992} proposed modular fixture systems that combine locators, clamps, and supports, while several groups developed pin-array devices~\cite{pinarray3,pinarray,pinarray2} that conform to object shapes under pressure. These rigid jigs enable high positioning accuracy but lack adaptability for variant–variety production.

To address this limitation, soft robotics technologies have been applied to jig design.
Brown~\etal~\cite{Brown2010} introduced a jamming gripper filled with granular particles that conforms to diverse object geometries. Inspired by this principle, jamming jigs have been proposed for robotic assembly tasks~\cite{Kiyokawa2020,Sakuma2021}, where deformable membranes adapt to shapes and fix objects after depressurization.
Kemmotsu~\etal~\cite{Kemmotsu2024} further developed a balloon pin-array gripper with two-step shape adaptation, combining axial sliding of pins and radial balloon inflation to achieve stable grasping even under misalignment; however, its dense pin arrangement may limit the reachable area after fixation, restricting its applicability to disassembly.
More recently, Aoyama~\etal~\cite{Aoyama2022} presented a shell-type hybrid gripper inspired by human fingers for compact packaging of agricultural products, and Hu~\etal~\cite{Hu2023} proposed a dual-mode enclosing soft gripper with tunable stiffness and high load capacity, capable of both contraction-based and suction-based grasping.

These works demonstrate the promise of adaptive soft jigs for general-purpose fixturing. Building on this direction, our study explores whether such deformable jigs can also function as effective regrasping tools in robotic disassembly tasks.
While regrasping studies have focused on planning and perception, and jig studies on adaptable fixation, 
few works have integrated soft, deformable jigs into regrasping frameworks. 
This gap motivates our investigation of jamming-inspired jigs specifically for regrasp-based uncertainty reduction.

\section{Placement Planning for Regrasping}
\subsection{Preconditions}
Regrasping of rigid components is performed using a robot arm equipped with a two-finger parallel gripper, a triangular-pyramid-shaped component placed within the arm’s workspace, and a flexible jig. 
To perform regrasping, we first determine the position and size of the cavity on the jig, as well as the object's dropping pose and final placement pose, the latter of which will be defined later based on the stability analysis. The triangular-pyramid-shaped object, grasped by the gripper, is then pressed onto the surface of the jig. While the object is being pressed, air is evacuated from the jig to induce the jamming transition, which increases the density of the internal granular material and solidifies the jig in the deformed shape.

After solidification, the triangular-pyramid object is returned to its original location, and a randomly oriented target part is picked up. This part is then dropped into the cavity formed in the soft membrane. The placement state of the part after dropping is recognized using an RGB-D camera. 
Finally, once the object is stably placed, it is regrasped and the system proceeds to the actual assembly task.

\subsection{Stamping Cavity Based on Feasibility and Stability}
During the shape formation process, the origin of the target coordinate frame $C_f$ for pressing the reference object is defined, as shown in~\figref{overview}(c). 
The transformation matrix $^{C_s}T_{C_f}$ from the source frame $C_s$ of the object to $C_f$ is computed, and a trajectory is generated accordingly. 
The frame $C_f$ is oriented such that all three axes form 45$^\circ$ angles with the horizontal plane.

Stability evaluation of each placement pose is performed with a two-stage criterion. 
First, we compute the wrench space at stamping depth $D$ using the simplified Ferrari--Canny method~\cite{Ferrari1992}:
\begin{equation} \forlab{convex-hull}
W_{L_\infty}(D) = \mathrm{ConvexHull}\Bigl(\Bigl\{\bigoplus_{i=1}^m \mathbf{w}_i \;\Big|\; \mathbf{w}_i \in W_i\Bigr\}\Bigr),
\end{equation}
where $\bigoplus$ denotes the Minkowski sum, $W_i$ is the set of wrenches at the $i$-th contact point, and $\mathbf{w}_i \in \mathbb{R}^{6}$ is the wrench (force and moment) generated at contact $i$. 
If the gravity wrench $\mathbf{w}_g$ lies inside $W_{L_\infty}(D)$, the pose is regarded as a stable placement pose (SPP). 
If this condition is not satisfied, we fall back to a geometric check: 
the convex hull $\mathcal{P}(D)=\mathrm{ConvexHull}(\{c_i(D)\}_{i=1}^m)$ of the contact points $c_i(D)$ projected onto the support plane is computed, 
and if the vertical projection of the COM $p_g$ lies inside $\mathcal{P}(D)$, the pose is also accepted as SPP. 
The pose obtained by vertically translating this SPP is defined as the deterministic dropping pose (DDP).

Based on this evaluation, we determine the stamping depth $D$ of the triangular-pyramid cavity that yields both a sufficient stability margin and the largest number of feasible grasp candidates~\cite{Wan2017}. 
The stamping depth $D$ further refines this geometry, since a deeper cavity generally increases stability but restricts approach directions, whereas a shallower cavity improves accessibility but reduces stability. 
Let $N_g(D)$ denote the number of feasible grasp candidates that admit at least one collision-free approach. 
The stability margin $M(D)$ is then defined as
\begin{equation}
    M(D) = 
    \begin{cases}
      \min\limits_{\mathbf{w} \in \partial W_{L_\infty}(D)} \|\mathbf{w}-\mathbf{w}_g\|_2, & \mathbf{w}_g \in W_{L_\infty}(D), \\[1ex]
      \mathrm{dist}(p_g, \partial \mathcal{P}(D)), & \text{otherwise},
    \end{cases}
\end{equation}
where $\mathrm{dist}(p_g, \partial \mathcal{P}(D))$ denotes the minimum distance from the COM projection to the boundary of the support polygon.

The optimal stamping depth $D^\star$ is then chosen as
\begin{equation}
    D^{\star} = \arg\max_{\,D \in [D_{\min},\,D_{\max}]} 
    \; \lambda N_g(D) + (1-\lambda) M(D),
\end{equation}
where $\lambda \in [0,1]$ is a weighting factor that trades off grasp feasibility against stability.
Intuitively, $M(D)$ measures how far the gravity wrench lies inside the feasible wrench polytope, while $N_g(D)$ counts how many grasp poses remain feasible; their weighted combination therefore balances physical stability and operational accessibility.

\figref{eval} illustrates the setup and analysis process used in this study. 
\figref{eval}~(a) shows the simplified simulation model employed for stability evaluation, 
where the cavity geometry and object were represented with sufficient fidelity while keeping computation tractable. 
\figref{eval}~(b) presents an example of candidate grasp poses generated on the object surface. 
\figref{eval}~(c) visualizes the stability analysis result, 
where the center of mass (red point) is projected onto the support plane and surrounded by the convex hull of the contact points (gray region). 
These analyses were used to determine whether a given pose can be regarded as a stable placement pose. 
\begin{figure}[tb]
  \centering
  \begin{minipage}[tb]{0.32\linewidth}
    \centering
    \includegraphics[width=\linewidth]{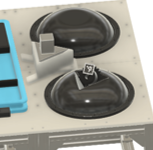}
    \subcaption{Simulation model}
  \end{minipage}
  \begin{minipage}[tb]{0.32\linewidth}
    \centering
    \includegraphics[width=\linewidth]{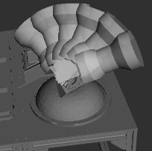}
    \subcaption{Grasp generation}
  \end{minipage}
  \begin{minipage}[tb]{0.32\linewidth}
    \centering
    \includegraphics[width=\linewidth]{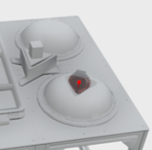}
    \subcaption{Stability analysis}
  \end{minipage}
  \caption{\small Examples of (a) simulation model for stability evaluation, (b) an example of generated grasps, and (c) stability analysis result showing COM and convex hull of contact points.}
  \figlab{eval}
\end{figure}

\section{Soft Jig for Triangular-Pyramid Cavities}
\begin{figure}[tb]
  \centering
  \includegraphics[width=\linewidth]{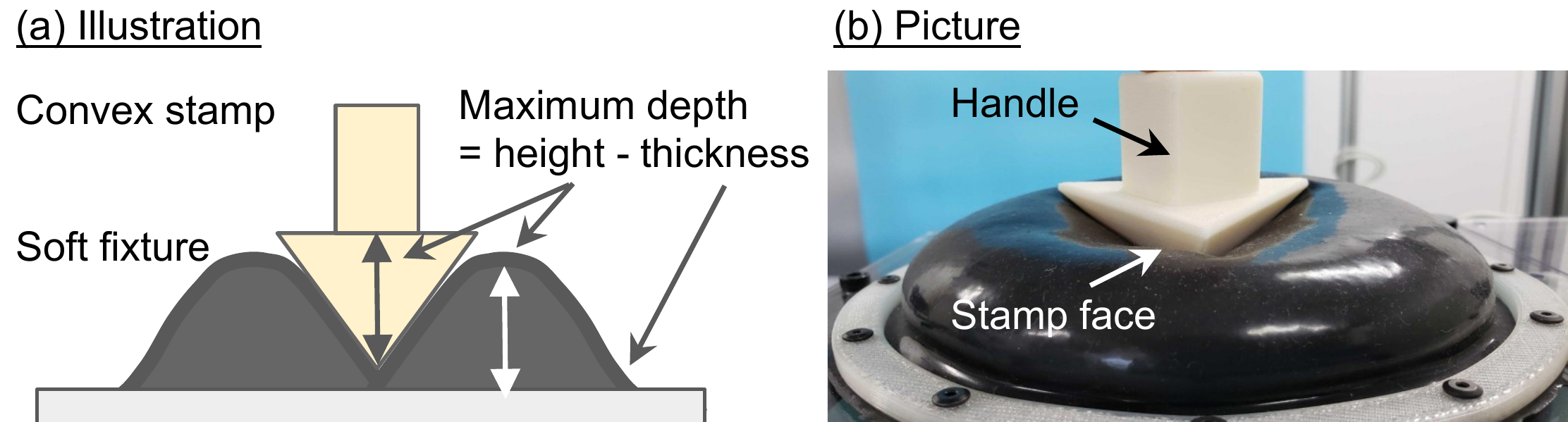}
  \caption{\small Geometries of a generated triangular-pyramid stamp and the resulting cavity.}
  \figlab{geometry}
\end{figure}
The cavity generation process involves fabricating a convex triangular-pyramid-shaped component with a handle, which is grasped by a robot arm and pressed against the flexible membrane. Subsequently, air is evacuated from the jig to form the desired cavity.
\figref{geometry} shows the scene where the convex triangular-pyramid component with a handle is pressed against the surface of the jig. The height of the triangular pyramid is designed to match the vertical thickness of the jig, including the membrane enclosing beads and air.

During suction, the density of the granular material inside the membrane increases, triggering the jamming transition, which transforms the internal state from fluid-like to solid-like. This results in a rigid, concave, triangular-pyramid-shaped cavity. Through this process, a firm contact surface between the placed object and the jig is formed.
The internal air pressure of the jig is controlled via a hose connected to the air inlet port on the jig. The hose is attached to a pump through an open/close valve, enabling binary ON/OFF control of the vacuum state by toggling the valve.

The jig design follows the soft-jig concept developed in previous studies~\cite{Kiyokawa2020}. 
The membrane is made of silicone rubber (1\,mm thickness, Shore A hardness of 2, Dragon Skin FX-Pro) to provide high elasticity, friction, and durability. 
The bag has a capacity of approximately 296\,cm$^3$ and is filled with 1\,mm glass beads, which are corrosion-resistant and enable repeatable jamming transitions. 
The curvature radius of the membrane surface is about 60\,mm, giving the jig sufficient deformability to generate cavities of different sizes. 
When vacuumed with an external pump, the internal pressure decreases by roughly 90\,kPa, stiffening the jig to hold objects rigidly. 
The initial shape of the membrane is reset by briefly applying positive pressure before stamping, since fixing performance depends strongly on the starting surface condition. 
The high friction coefficient of the silicone membrane (measured around 1.97 in sliding tests~\cite{Kiyokawa2020}) also contributes to stable contact with target parts. 
These design choices, determined through preliminary trials, provided the best balance between deformability and robustness: 
the 1\,mm membrane maximized deformation without tearing, 
while the 1\,mm glass beads offered a compromise between flowability during evacuation and rigidity after jamming. 
Overall, the configuration grants the jig the versatility to accommodate different object geometries while maintaining sufficient rigidity during regrasping.

\section{Performance Evaluation}
\begin{figure}[tb]
  \centering
  \includegraphics[width=\linewidth]{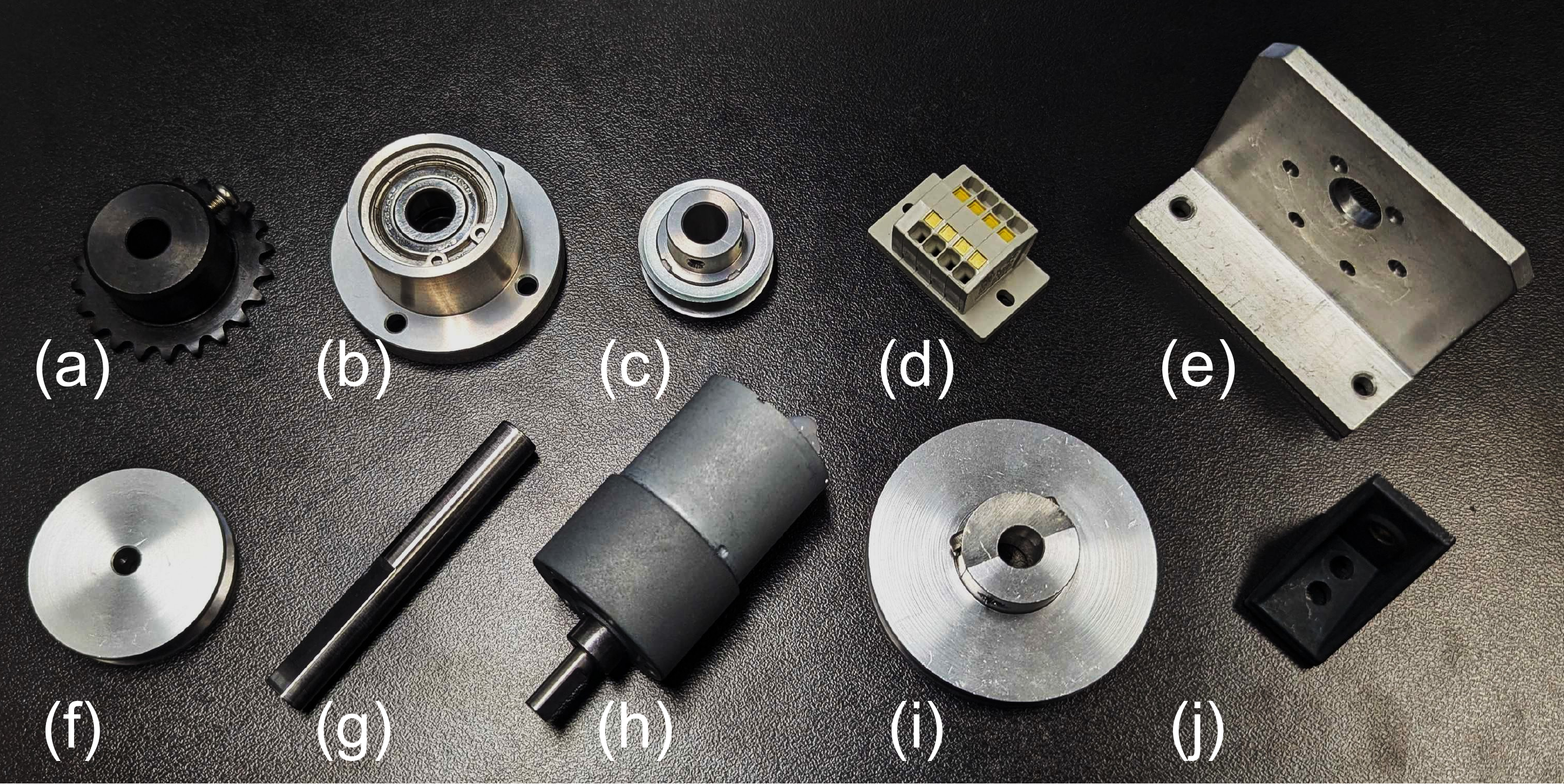}
  \caption{\small Target objects used for our experiments.}
  \figlab{object}
\end{figure}
\subsection{Overview}
To evaluate the proposed method for jig-assisted regrasping, we conducted two experiments: (1) to measure the deviation between the triangular-pyramid-shaped cavity generated by pressing a convex shape and the ideal target shape, and (2) to assess the success rate and repeatability of part-dropping trials.
As shown in~\figref{overview}, the experimental setup consists of the developed flexible jig, convex triangular-pyramid-shaped components fabricated with a 3D printer for generating the cavity on the jig surface, and a robot arm equipped with a two-finger gripper.
For Experiment (1), we additionally use an Intel RealSense D455 depth camera to acquire 3D information.
For Experiment (2), the weighting factor defined in Equation~(3) was empirically set to $\lambda=0.5$, which provided a good balance between stability and grasp feasibility in preliminary trials, and this value was used for all subsequent experiments.
Sensitivity checks with $\lambda\in[0.3,0.7]$ produced the same qualitative trends, indicating that the optimization is not overly sensitive to moderate changes in $\lambda$.

\figref{object} shows the ten different types of target objects used in the experiments. 
These include common mechanical components such as sprockets, pulleys, brackets, and shafts, which must be precisely assembled in actual tasks. 
\tabref{object} provides detailed information about these objects, including type, mass, and dimensions. 
All ten objects were used in both experiments to verify the versatility of the proposed jig.
\begin{table}[tb]
    \small
    \centering
    \caption{\small{Target Object Information}}
    \begin{threeparttable}
        \begin{tabular}{llrr} \toprule
            \multicolumn{1}{c}{Object ID} & \multicolumn{1}{c}{Type} & \multicolumn{1}{c}{Mass [g]} & \multicolumn{1}{c}{WxLxH [mm]} \\ \midrule
            (a) & Sprocket & 106 & 48x48x18 \\
            (b) & Bearing Holder & 119 & 54x54x30 \\
            (c) & Timing Pulley & 19 & 32x32x20 \\
            (d) & Terminal Block & 12 & 23x39x20 \\
            (e) & L-Bracket for Motor & 75 & 25x70x60 \\
            (f) & Idler Pulley & 32 & 42x42x10 \\
            (g) & Shaft & 45 & 10x10x75 \\
            (h) & Geared DC Motor & 190 & 37x82x37 \\
            (i) & Round Belt Pulley & 84 & 62x62x20 \\
            (j) & Small L-Bracket & 14 & 20x30x30 \\ \bottomrule
        \end{tabular}
        \vspace{0.5mm}
    \end{threeparttable}
    \tablab{object}
\end{table}
\begin{figure}[tb]
  \centering
  \begin{minipage}[tb]{0.495\linewidth}
    \centering
    \includegraphics[width=\linewidth]{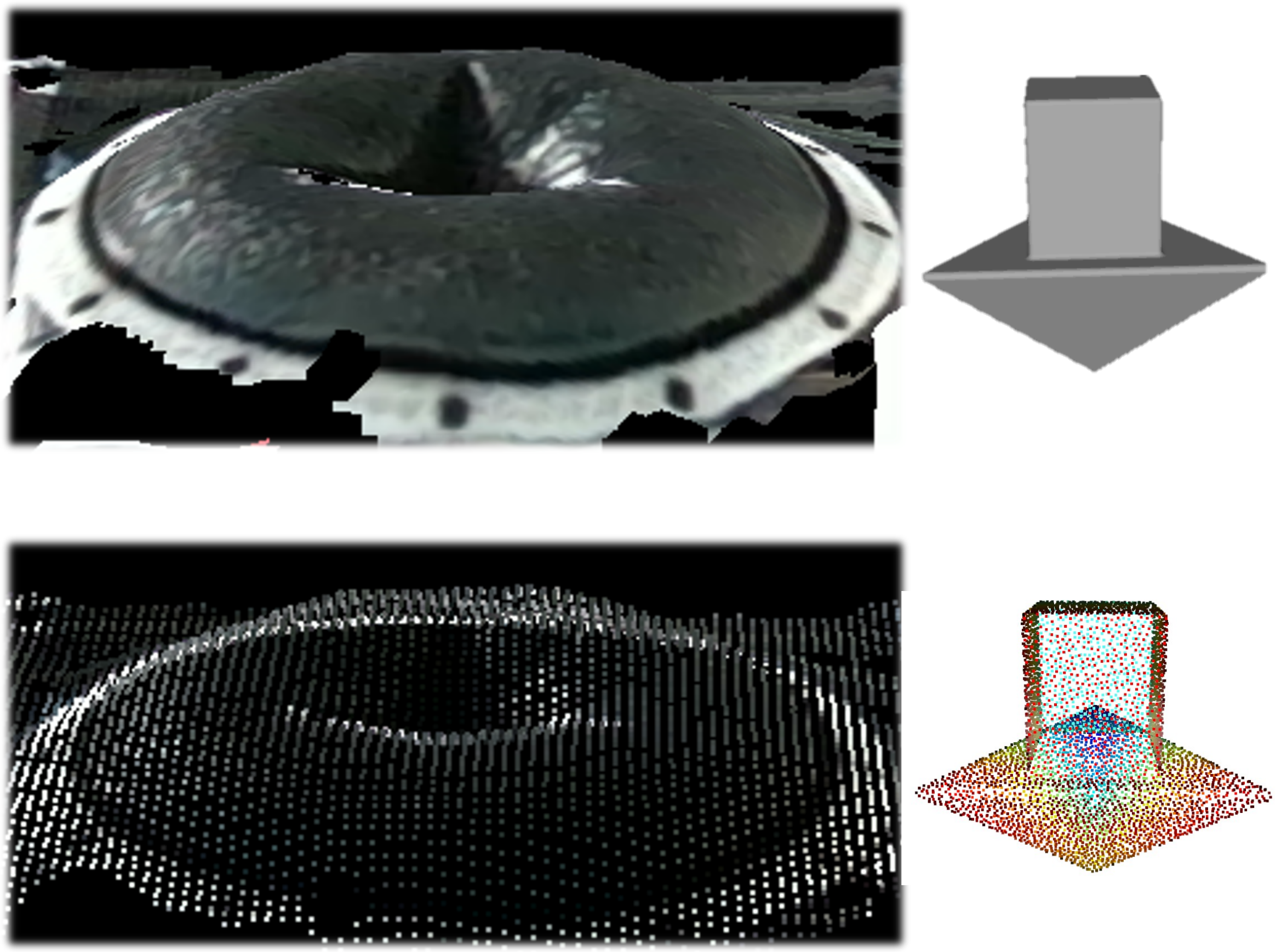}
    \subcaption{Meshes and point clouds}
  \end{minipage}
  \begin{minipage}[tb]{0.47\linewidth}
    \centering
    \includegraphics[width=\linewidth]{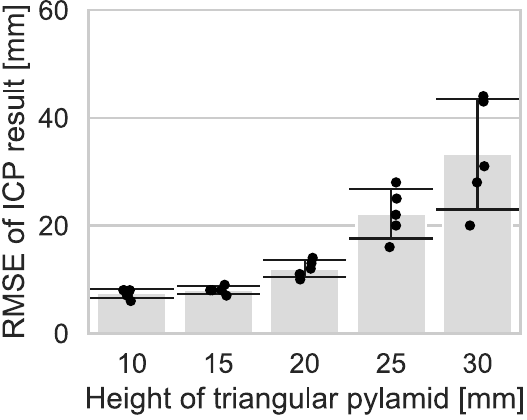}
    \subcaption{Shape errors}
  \end{minipage}
  \caption{\small Comparison of the target and generated shapes}
  \figlab{gen-result}
\end{figure}

\subsection{Generated Triangular-Pyramid Shape}
To quantify the difference between the generated triangular-pyramid shape and the ideal target shape, depth information is acquired using a depth sensor. In this experiment, the shape generation performance of the flexible jig is evaluated by calculating the geometric error between the generated shape and the target triangular-pyramid shape planned by the proposed method.
Specifically, point cloud data is extracted from the depth images obtained by the sensor, within the contour of the generated triangular-pyramid-shaped region. This point cloud is then registered to a point cloud representation of the target triangular-pyramid shape using RANSAC (RANdom SAmple Consensus) followed by ICP (Iterative Closest Point). The registration residual error is used as the metric for evaluation.

\figref{gen-result}~(a) shows mesh (top row) and point cloud (bottom row) representations of both the generated shape (left column) and the target shape (right column). Qualitatively, the shapes are highly similar, and even manual object placement trials indicated that objects readily settled into the bottom of the cavity.
\figref{gen-result}~(b) shows the quantitative shape errors. The root mean square error (RMSE) after point cloud registration was at most 4.4~[mm], indicating that the deviation in surface geometry was sufficiently small.
Considering that the target parts have characteristic dimensions in the range of 30--80\,mm, this maximum error of 4.4\,mm is relatively small and did not noticeably affect the ability of the objects to settle into the cavity or the overall placement accuracy.
This error corresponds to less than 6\% of the typical part size, and manual tests confirmed that objects could still settle stably.

To prevent damage to the flexible membrane, the edges of the pressing component were filleted during 3D printing. As a result, the generated concave edges were smoother than those of rigid jigs used in previous studies~\cite{Hu2024}. This difference in edge sharpness sometimes caused minor variations in the object’s final resting position. Improving the membrane's durability may lead to further improvements in precision.

\subsection{Part-Dropping Accuracy}
For all ten objects, the optimization in Equation~(3) was successfully applied without failure to all ten objects, demonstrating robustness of the optimization pipeline across diverse geometries.
Using the SPP--DDP pairs planned by the proposed method, drop experiments were conducted to evaluate object placement accuracy with cavities of the corresponding planned size. 
All ten objects were tested, and for each object 20 drop trials were performed. 
A trial was regarded as successful when the position and orientation errors detected by the visual markers were within 5\,mm and 5$^\circ$, respectively. 

The vertical drop height was set to twice the maximum depth of the cavity (\figref{geometry}), a value empirically found to minimize failed placements during preliminary testing. 
If the drop height is too small, parts may contact the jig edge prematurely due to pose uncertainty, leading to unstable or biased placements. 
Conversely, an excessively large drop height increases the impact at contact and often causes bouncing because of the elasticity of the silicone membrane. 
Setting the height to approximately twice the cavity depth provided a balance between these effects,  allowing parts to settle reliably at the cavity bottom.

\begin{figure}[tb]
  \centering
  \includegraphics[width=\linewidth]{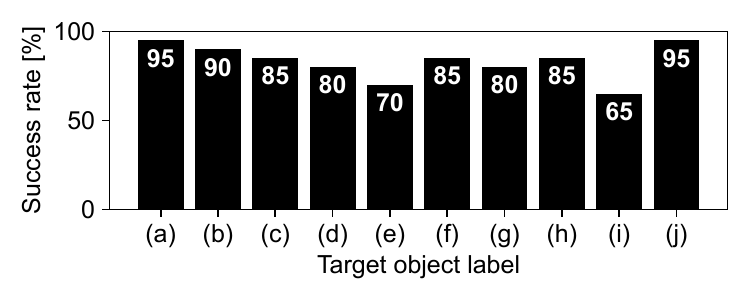}
  \caption{\small Placement performance. X-axis label corresponds to the object labels assigned in~\figref{object}.}
  \figlab{performance}
\end{figure}
\figref{performance} shows the success rates [\%] obtained in the experiments. 
Overall, the proposed method achieved high success rates, with most objects exceeding 80\%. 
In particular, objects (a), (b), and (j) reached 90\% or higher. 
In contrast, objects (e), (g), (h), and (i) exhibited lower success rates of 65--80\%. 

\begin{figure}[tb]
  \centering
  \begin{minipage}[tb]{0.48\linewidth}
    \centering
    \includegraphics[width=\linewidth]{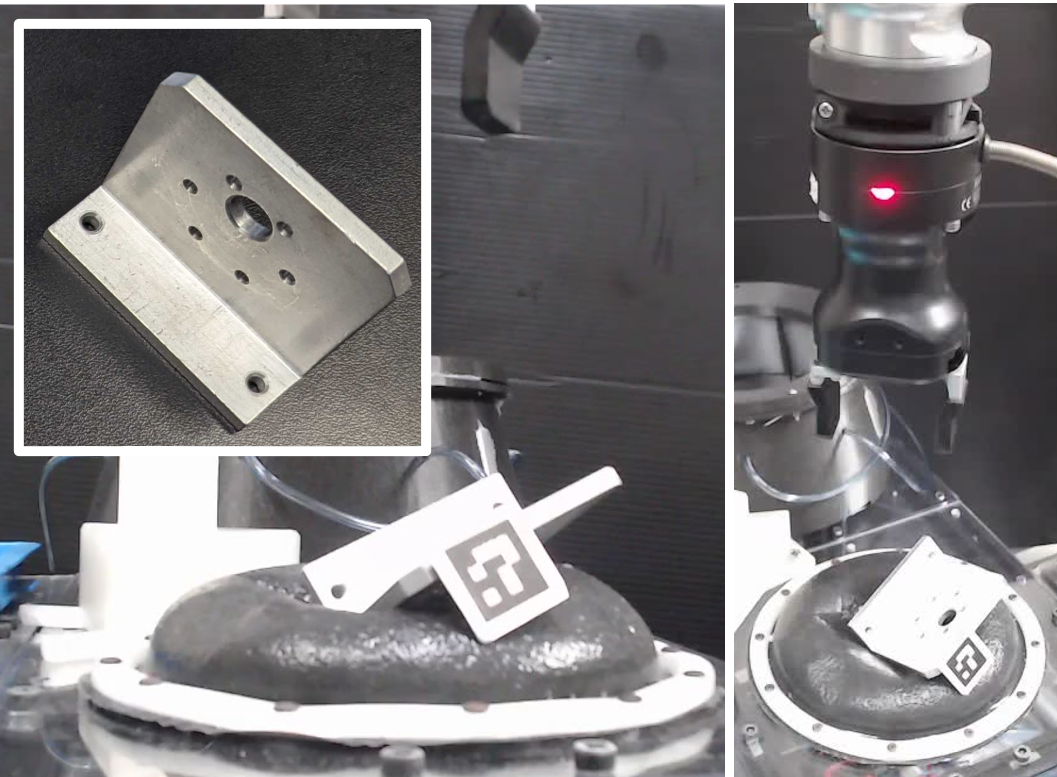}
    \subcaption{Object (e)}
  \end{minipage}
  \begin{minipage}[tb]{0.48\linewidth}
    \centering
    \includegraphics[width=\linewidth]{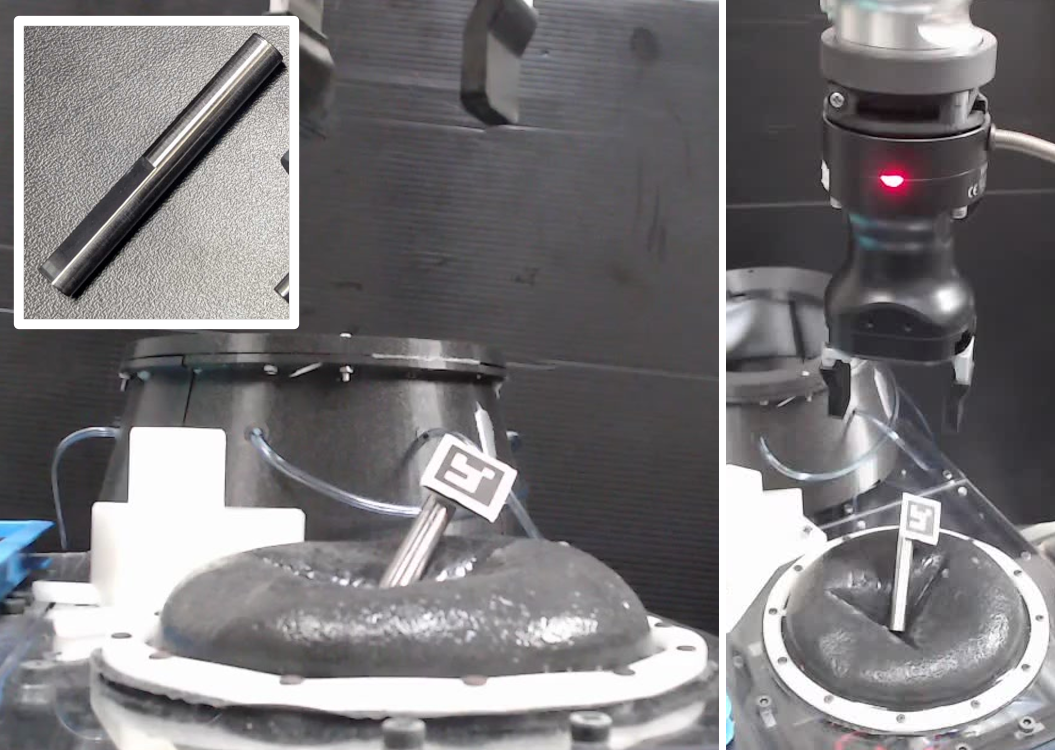}
    \subcaption{Object (g)}
  \end{minipage}
  \begin{minipage}[tb]{0.48\linewidth}
    \centering
    \includegraphics[width=\linewidth]{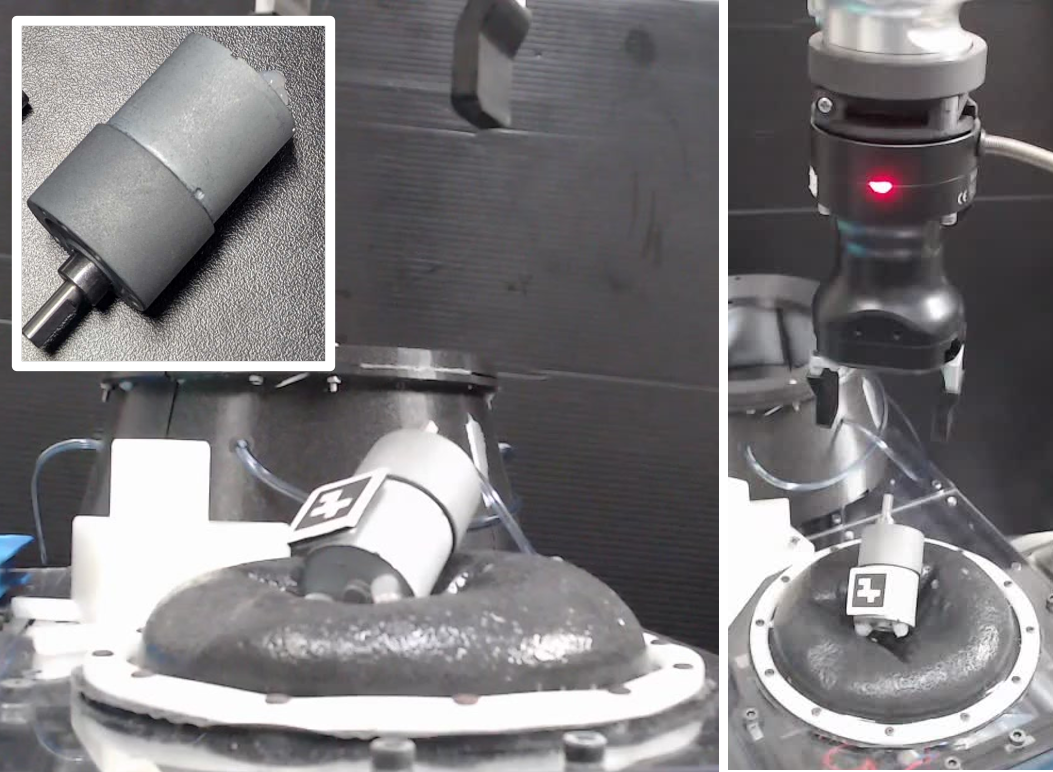}
    \subcaption{Object (h)}
  \end{minipage}
  \begin{minipage}[tb]{0.48\linewidth}
    \centering
    \includegraphics[width=\linewidth]{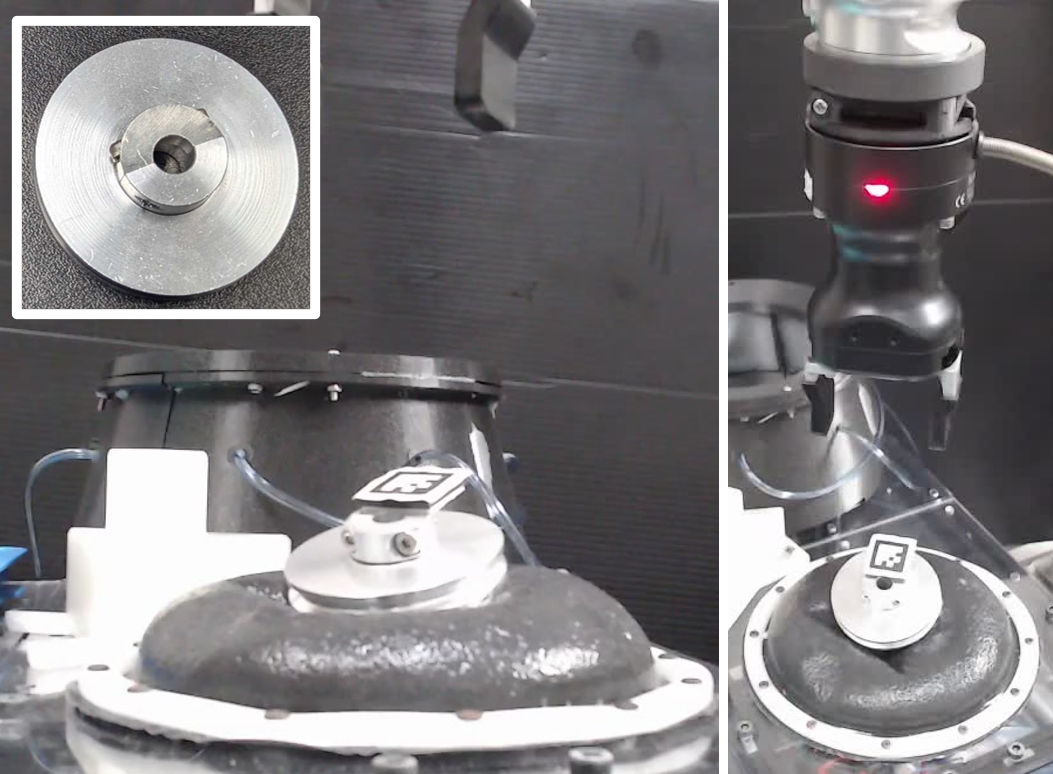}
    \subcaption{Object (i)}
  \end{minipage}
  \caption{\small Final placement states for objects (e), (g), (h), and (i). Even in successful cases, the objects were only marginally stable, often resting at tilted or precarious angles.}
  \figlab{difficult}
\end{figure}

\figref{difficult} shows the final placement states after dropping for objects (e), (g), (h), and (i). 
These cases illustrate the inherent difficulty of parts with sloped or asymmetric surfaces: geometric interference prevented the objects from fully reaching the cavity bottom, so even when a trial was counted as “success,” the resulting placement was often barely stable. 
For the remaining cylindrical parts, most of the failed placements were instead caused by bouncing upon impact or insufficient sliding into the cavity, both of which are linked to the elasticity and frictional properties of the silicone membrane.

These results suggest that while complex-shaped parts (e.g., pulleys or bearing holders) remain challenging, the proposed flexible jig system provides stable and repeatable regrasping performance for a wide variety of small mechanical components.

\section{Discussion}
During the experiments, small and lightweight parts often failed to settle reliably at the bottom of the cavity. 
Even when they entered the cavity, their final poses sometimes deviated significantly from the desired configuration. 
These issues were mainly caused by the elasticity of the silicone membrane, which induced bouncing, and the high surface friction, which hindered smooth sliding. 

In future deployments, simple measures such as surface lubrication and external vibration could be incorporated to mitigate these effects. 
As a preliminary check, we tested these interventions on a subset of parts and observed that the overall success rate increased by roughly 5\%, 
indicating that such refinements have clear potential to enhance the proposed method.

\begin{figure}[tb]
  \centering
  \begin{minipage}[tb]{0.32\linewidth}
    \centering
    \includegraphics[width=\linewidth]{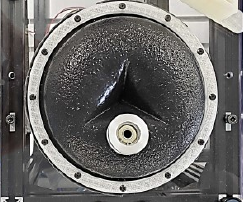}
    \subcaption{Failure}
  \end{minipage}
  \begin{minipage}[tb]{0.32\linewidth}
    \centering
    \includegraphics[width=\linewidth]{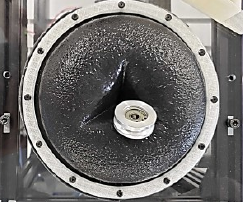}
    \subcaption{Near-miss}
  \end{minipage}
  \begin{minipage}[tb]{0.32\linewidth}
    \centering
    \includegraphics[width=\linewidth]{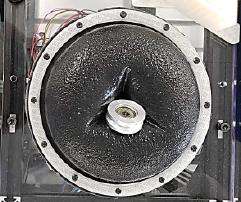}
    \subcaption{Success}
  \end{minipage}
  \begin{minipage}[tb]{\linewidth}
    \centering
    \includegraphics[width=\linewidth]{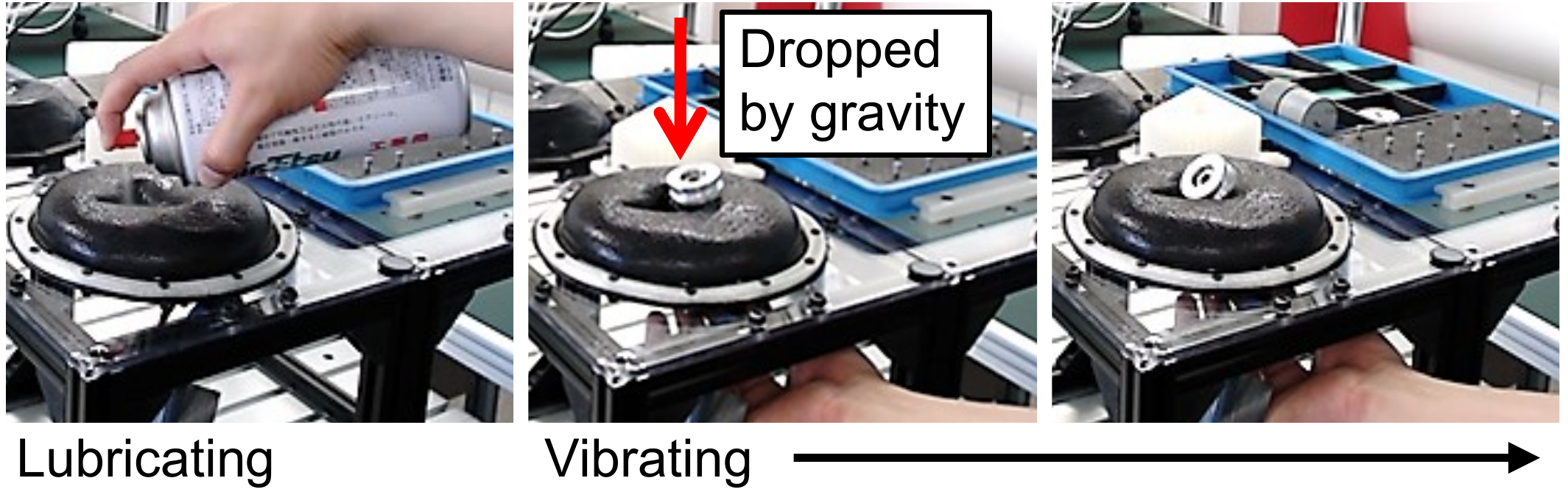}
    \subcaption{Process to derive the part to slide to the bottom}
  \end{minipage}
  \caption{\small Failure and success cases.}
  \figlab{cases}
\end{figure}
\figref{cases} illustrates representative outcomes for another cylindrical shape, including failed placements (Failure), near-miss cases (Near-miss), and successful placements (Success). 
Failures were generally of two types: (i) the part failed to slide at all and stopped near the cavity edge, and (ii) the part slid partially but stopped midway. 

In the former case, recovery through vibration was difficult because the part did not engage with the cavity surface. 
In the latter case, however, the part remained in partial contact with the slope, and vibration was able to induce successful placement. 
In successful cases, the part reached the cavity bottom in a manner similar to rigid triangular-pyramid jigs, enabling reliable regrasping. 

\begin{figure}[tb]
  \centering
  \begin{minipage}[tb]{0.45\linewidth}
    \centering
    \includegraphics[width=\linewidth]{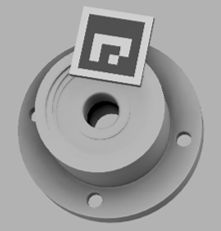}
    \subcaption{AR marker attached}
  \end{minipage}
  \begin{minipage}[tb]{0.528\linewidth}
    \centering
    \includegraphics[width=\linewidth]{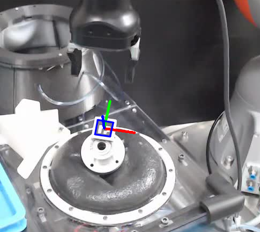}
    \subcaption{Detected AR marker}
  \end{minipage}
  \caption{\small AR markers used in the dropping experiments.}
  \figlab{marker}
\end{figure}
Another limitation arises from the experimental setup. 
To measure object pose, a lightweight AR marker plate (2\,g, 25\,mm $\times$ 25\,mm $\times$ 2\,mm) was attached to each target part, as illustrated in \figref{marker}. 
Although the added mass was small (2\,g), the offset created torque shifts noticeable for lightweight parts, occasionally altering their settling orientation compared to marker-free conditions.
Although the marker offset was modeled when estimating reachability and stamping depth, the additional plate inevitably shifted the center of mass, producing mass distributions that differ from the actual components. 
This discrepancy represents a limitation of the current evaluation.

\section{Conclusion}
This study presented a regrasp planning and execution framework using a deformable soft jig with jamming transition. 
We demonstrated how triangular-pyramid-shaped cavities can be generated on the jig surface and introduced a depth--approachability optimization to balance placement stability and grasp feasibility. 

Experiments with ten diverse parts confirmed high success rates, particularly for cylindrical objects, while lightweight and geometrically constrained parts remained challenging due to bouncing, friction, or entrapment on sloped surfaces. 
Although simple interventions such as lubrication and vibration improved results, these measures are ad hoc, and systematic solutions are required. 
Another limitation arose from the AR markers used for pose detection, which shifted the center of mass and may not fully reflect actual part behavior.

Future work will address these limitations and extend jig-assisted assembly planning. 
We aim to integrate flexible jigs into sequence optimization frameworks that consider part-to-part and part-to-jig relationships~\cite{Tariki2020ar,Kiyokawa2021seq}, combine them with pose estimation~\cite{Sakuma2022}, and explore adaptive generation of convex triangular-pyramid pressing tools whose size and geometry are selected according to stability analysis of target parts. 
Such directions will further improve the generality and robustness of flexible regrasping for assembly automation.

\bibliographystyle{IEEEtran}
\footnotesize
\bibliography{reference}

\end{document}